\documentclass[runningheads]{llncs}
\usepackage{wrapfig}
\usepackage{multirow}
\usepackage{amsmath}
\usepackage{amssymb}
\usepackage{makecell}

\usepackage[final,year=2026,ID=461]{accv}

\usepackage{accvabbrv}

\usepackage{graphicx}
\usepackage{booktabs}
\renewcommand{\arraystretch}{0.95}
\usepackage[accsupp]{axessibility}  

\usepackage[pagebackref,breaklinks,colorlinks,citecolor=accvblue]{hyperref}

\usepackage{orcidlink}

\begin{document}

\title{Weave Mamba Fusion: Global Cross-Scale Interaction for Lightweight Face Detection}
\titlerunning{Weave Mamba Fusion}

\author{
    Dohun Kim\inst{1}\orcidlink{0009-0003-5219-0715}, 
    Jinmyung Jung\textsuperscript{\dag}\orcidlink{0000-0002-4962-8347}
}

\institute{
    School of Data Science, The University of Suwon, Korea \\
    \email{dohun0714@naver.com, jmjung@suwon.ac.kr}
    \\[0.5em]
    \textsuperscript{\dag}Corresponding Author
}

\authorrunning{D.~Kim et al.}



\maketitle

\begin{abstract}
Feature pyramid methods, from FPN to BiFPN, have achieved strong performance in
face detection by fusing multi-scale features. However, detecting faces under
unconstrained conditions, such as small scale, occlusion, and extreme pose,
remains difficult, as it requires global cross-scale dependencies that
local fusion cannot model. State space models such as Mamba provide global context with linear complexity by scanning features as a sequence, and therefore offer a promising direction for this problem. Nevertheless, such a scan needs the two pyramid scales combined into a single feature map, and the way they are combined determines whether cross-scale structure is preserved. Summation collapses the two scales before the scan, so the scan has no cross-scale structure to exploit, while concatenation keeps both scales but at far higher cost. To address this, we propose
\textbf{Weave Mamba Fusion (WMF)}, which interleaves two adjacent pyramid scales column by column so that each step of a horizontal bidirectional SS2D scan moves from one scale to the other. With partial-channel processing and parameter-free de-weaving, WMF enables efficient cross-scale interaction while preserving feature structure. Integrating WMF into every fusion node yields
\textbf{WeaveBiFPN}, the neck of our \textbf{WeaveFace} detector. On WIDER FACE, WeaveFace achieves 91.41\% mean AP with only 0.34M parameters and 1.16 GFLOPs, outperforming prior detectors under 0.5M parameters. Its largest gains are on the Hard subset, where it reaches 87.14\% AP.
The code is publicly available at \url{https://github.com/dohun-mat/WeaveMambaFusion}.

\vspace{1em} 
\end{abstract}

\section{Introduction}
\begin{figure*}[t]
    \centering
    \includegraphics[width=\linewidth]{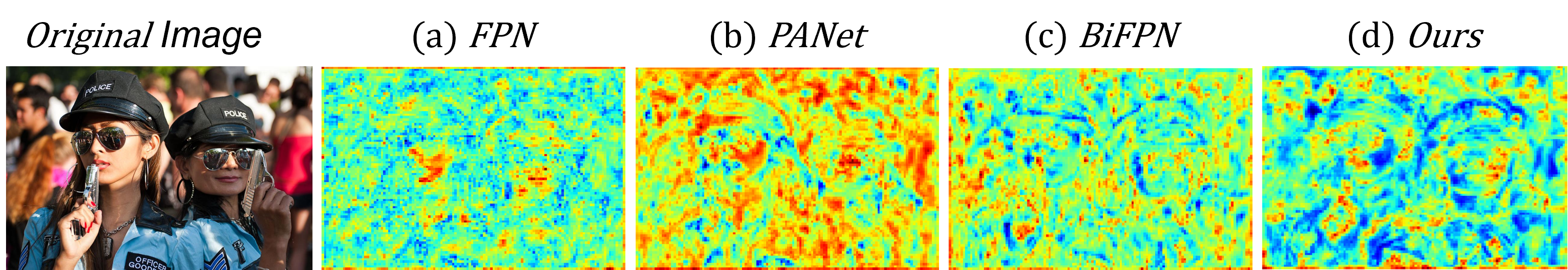}
    \caption{Feature map visualization of different feature pyramid networks. Given the input image, we visualize the fused features from (a) FPN, (b) PANet, (c) BiFPN, and (d) our WeaveBiFPN. Our WeaveBiFPN separates faces from the background more clearly than the others.}
\label{fig:featuremap}
\end{figure*}

Face detection enables a wide range of applications such as face
recognition, biometric authentication, and human--computer interaction.
Despite substantial progress driven by deep convolutional networks,
detection in unconstrained, real-world environments remains an open
problem. Challenges such as extreme scale variation, severe occlusion,
pose variation, and blur are most pronounced on the \emph{Hard} subset
of the WIDER FACE benchmark~\cite{widerface}.

\begin{figure}[t]
    \centering
    \includegraphics[width=0.8\linewidth]{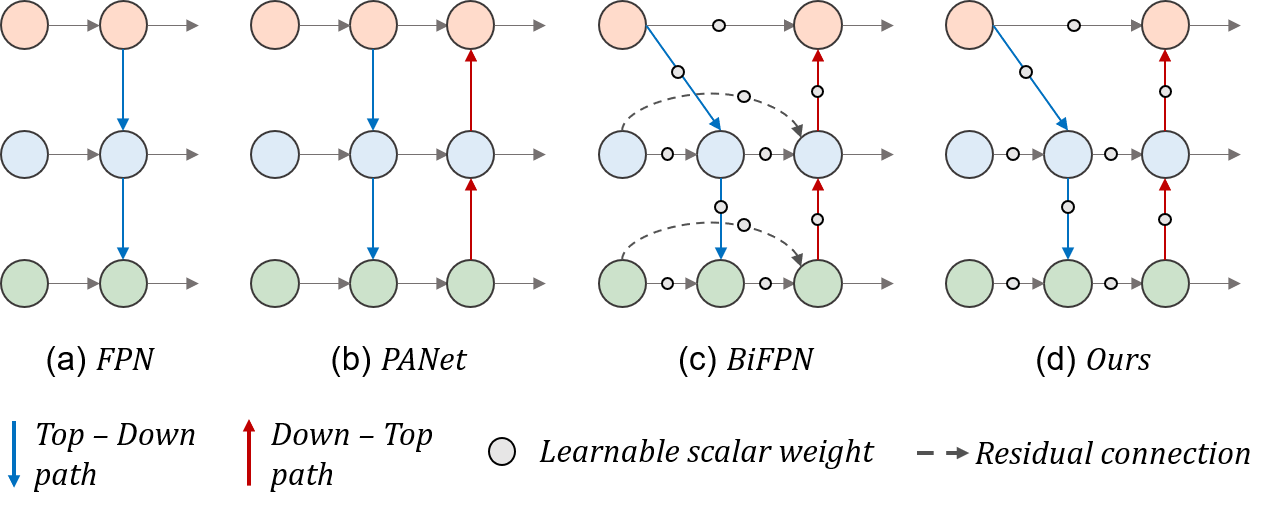}
    \caption{Comparison of feature pyramid architectures. (a) FPN. (b) PANet. (c) BiFPN. (d) Our WeaveBiFPN. The dashed boxes enlarge the feature fusion in each design. Our WeaveBiFPN keeps the bidirectional structure of BiFPN but replaces its weighted fusion with Weave Mamba Fusion.}
\label{fig:neck_comparison}
\end{figure}

Reliable detection under such conditions requires combining the
fine-grained spatial information of high-resolution features with the
rich semantics of low-resolution ones, making multi-scale feature fusion
an essential ingredient of modern detectors. FPN~\cite{fpn}, PANet~\cite{panet}, and BiFPN~\cite{bifpn} have progressively
refined how information flows across pyramid levels---adding bottom-up
paths and learnable fusion weights atop FPN's original top-down design
(Fig.~\ref{fig:neck_comparison}(a)--(c)). Yet even with learnable weights, their cross-scale fusion provides only local interaction---summation or concatenation~\cite{yolov4} followed by local convolutions---and cannot reach distant context.

As shown in Fig.~\ref{fig:featuremap}, the fused features therefore lack global context, a limitation most evident for small and occluded faces.  
Such faces provide little information on their own, so detecting them relies on surrounding semantic context. This context lies in the low-resolution pyramid level, but at positions away from the face, not at the face's own location. A local fusion can only combine features at nearby positions and thus cannot reach this distant context. Detecting these faces therefore calls for \emph{global, cross-scale}
interaction in which each position can integrate distant context from the
other scale.

Overcoming this limitation requires a fusion module that (i) retains the
spatial information from both scales, (ii) enables long-range, cross-scale
interaction rather than the spatially local interaction inherent to convolution,
and (iii) remains efficient enough to operate at every node of a high-resolution
pyramid. Self-attention~\cite{attention} provides all-pairs interaction—one way to achieve (ii)—but its quadratic complexity becomes prohibitively expensive at the high resolutions where pyramid fusion operates. State space models (SSMs) such as Mamba~\cite{mamba}, by contrast, offer global context at linear complexity. An SSM alone, however, is not enough. Once the two scales are collapsed by summation, the scan has no cross-scale structure to exploit. The key insight is that, because a scan is sequential, \emph{weaving} the two scales lets a single scan carry information from one scale into the other at every step.

Motivated by this, we propose \textbf{Weave Mamba Fusion (WMF)}, a fusion
operator that achieves global cross-scale interaction at linear cost. WMF
weaves the two feature maps column by column into a single 2D map and
processes it with a bidirectional horizontal scan built on
SS2D~\cite{visionmamba,vmamba}. Because adjacent columns belong to
different scales, every step of the scan propagates information from one
scale to the other, giving each location access to the full spatial extent of the other scale rather than a local neighborhood. To keep this efficient, we further adopt a partial-channel design inspired
by partial convolution (PConv)~\cite{fasternet}, applying SS2D to only a
fraction of the channels and leaving the rest as an identity branch, which substantially reduces the cost of the scan with negligible loss in fusion quality.
Applying WMF at every cross-scale node of both BiFPN pathways yields
\textbf{WeaveBiFPN}, the neck of our \textbf{WeaveFace} detector, which
attains state-of-the-art accuracy among lightweight face detectors,
with its largest gain on the \emph{Hard} subset where long-range
cross-scale context matters most.

The main contributions of this paper are summarized as follows:
\begin{itemize}
    \item We propose \textbf{Weave Mamba Fusion (WMF)}, an SSM-based
    redesign of the multi-scale fusion operator for face detection. Rather
    than summation or concatenation followed by local convolutions, WMF
    weaves two feature maps into a single 2D map and processes it with a
    bidirectional scan, achieving global cross-scale interaction within a
    single operator.
    \item We introduce a partial-channel design that applies the scan to only a fraction of the channels, substantially reducing WMF's dominant cost while preserving cross-scale fusion.
    \item We propose \textbf{WeaveBiFPN} and the resulting
    \textbf{WeaveFace} detector. On WIDER FACE, WeaveFace attains 87.14\%
    Hard AP and 91.41\% mean AP with only 0.34M parameters and 1.16
    GFLOPs, making it the most accurate detector with fewer than 0.5M
    parameters.
\end{itemize}

\section{Related Work}
\subsection{Face Detection}
Driven by advances in deep CNNs, face detection has progressed rapidly. Early
work such as MTCNN~\cite{mtcnn} learns face detection and landmark localization
together through cascaded multi-task learning. RetinaFace~\cite{retinaface}
adopts a multi-branch prediction head and has become a widely used baseline for
face detection in unconstrained environments. DSFD~\cite{dsfd} improves
small-face detection through dual-shot detection and feature enhancement, and
TinaFace~\cite{tinaface} shows that competitive face detection accuracy can be
achieved using only generic object detection techniques. These detectors~\cite{retinaface, dsfd, tinaface, mogface, efficientsrface},
however, achieve such accuracy only at the cost of tens of millions of  parameters and tens to hundreds of GFLOPs.

Lightweight face detection has also been actively studied. SCRFD~\cite{scrfd}
improves efficiency by redistributing training samples and computation toward
the smaller, harder faces. EResFD~\cite{eresfd} shows that a ResNet with greatly
reduced channels outperforms previous lightweight backbones, and
FDLite~\cite{fdlite} reduces the parameter count to only 0.24M. More recent
detectors instead raise accuracy by strengthening the BiFPN neck, as in
FeatherFace~\cite{featherface}, which adds CBAM attention and deformable
convolutions. 

\subsection{Feature Pyramid Networks}
FPN~\cite{fpn, wang2017face, pyramidbox} introduces a multi-scale fusion
architecture in which a top-down pathway passes semantic information from
deeper levels to shallower ones, while lateral connections add the matching backbone features at each level. PANet~\cite{panet} adds a bottom-up pathway to reinforce high-level features with low-level spatial information, and BiFPN~\cite{bifpn} uses a bidirectional cross-scale structure with learnable fusion weights. Beyond these hand-designed pathways, AFPN~\cite{afpn} fuses adjacent levels progressively before incorporating distant ones.

Despite their different topologies, these methods share the same fusion
operator: two levels are resized to a common spatial size and merged
position-wise, followed by convolutions with a local receptive field. Their cross-scale interaction is therefore confined to a local neighborhood and cannot capture long-range dependencies to distant context in the other scale. FPT~\cite{fpt} instead models non-local interaction via attention across pyramid levels, but its quadratic cost makes it ill-suited to the lightweight regime. We instead replace the local, position-wise fusion operator with Weave Mamba Fusion (WMF), which achieves global cross-scale interaction at linear cost.

\subsection{State Space Models for Multi-Scale Fusion}
State space models (SSMs) have gained attention as an alternative to
Transformers, modeling long-range dependencies with linear complexity in
sequence length. The Structured State Space sequence model (S4)~\cite{s4}
established the efficiency of long-range sequence modeling, and
Mamba~\cite{mamba} advanced it with a selective SSM (S6) whose parameters
depend on the input. Vision Mamba~\cite{visionmamba} and VMamba~\cite{vmamba}
extend Mamba to the image domain; in particular, VMamba introduces the Selective Scan 2D (SS2D), which combines scans along multiple directions to
model global information in 2D images.

Within detection, most SSM-based methods place the scan in the
\emph{backbone}~\cite{mambayolo,mambavision,wmamba}, extracting single-scale
features before any cross-scale fusion occurs. A smaller body of work
instead places SSMs in the \emph{fusion pathway}: Mamba-YOLO's~\cite{mambayolo}
PAFPN neck, PyramidMamba~\cite{pyramidmamba}, and MDDCNet~\cite{mddcnet} all
add SSM blocks alongside multi-scale connections, but the cross-scale
combination itself remains a conventional position-wise operation, with the
SSM only enhancing features before or after it. WMF instead applies the SSM
\emph{as} the fusion operator: the two scales are woven at alternating
positions, so every step of the SS2D scan moves from one scale to the other.

\section{WeaveFace}
\label{sec:method}
\begin{figure}
    \centering
    \includegraphics[width=1\linewidth]{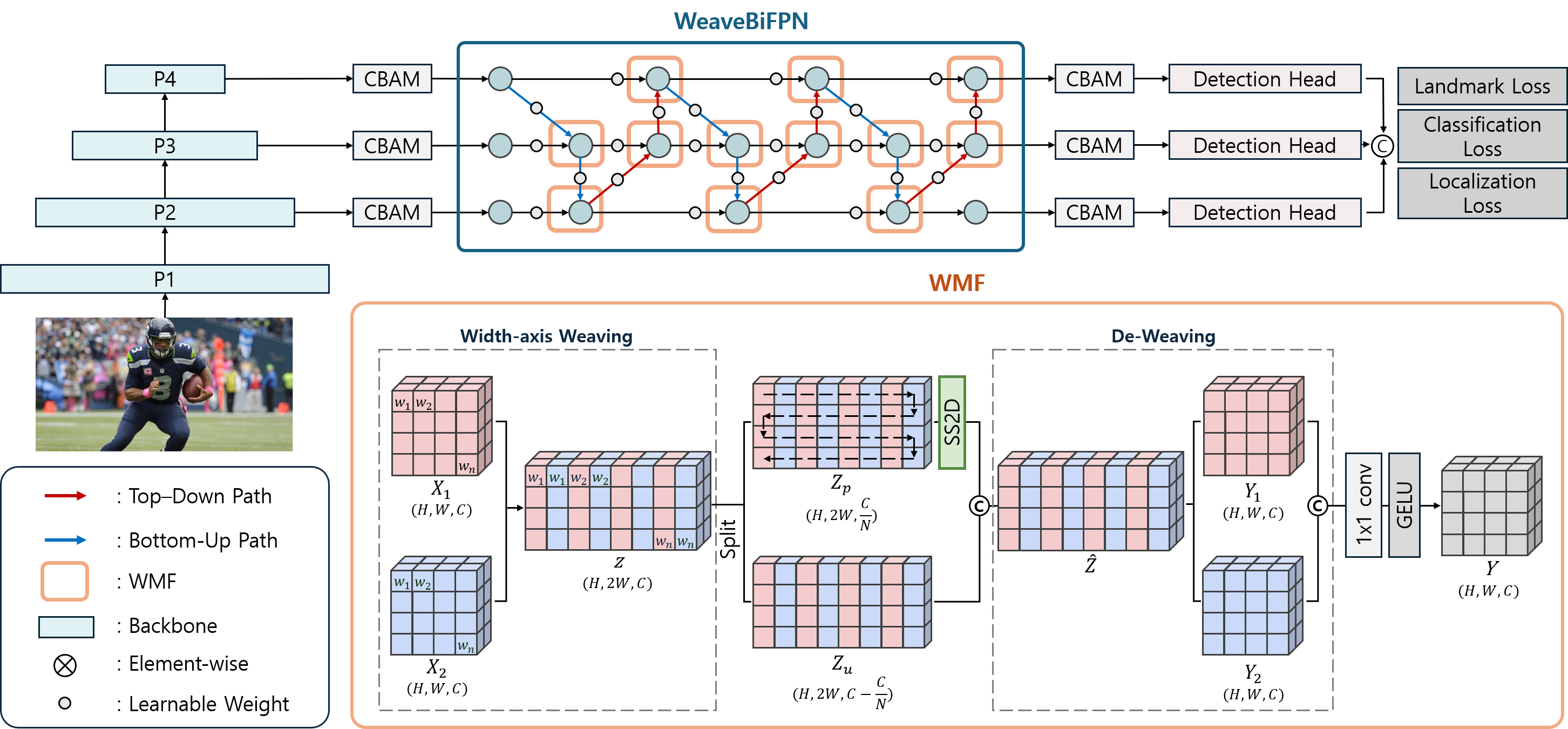}
    \caption{Overview of \textbf{WeaveFace}.}
    \label{fig:overview}
\end{figure}

\label{sec:overview}
The proposed WeaveFace architecture, illustrated in
Fig.~\ref{fig:overview}, follows a standard detection pipeline,
comprising an EResNet backbone~\cite{eresfd} and CBAM~\cite{cbam}
refinement modules. Our core contribution lies in the WeaveBiFPN neck, which integrates our novel Weave Mamba Fusion (WMF) to enable global cross-scale feature interaction. The fused features are passed to task-specific detection heads to produce classification, bounding box, and landmark outputs. These heads are optimized using a standard multi-task loss, which effectively leverages the global cross-scale features captured by WMF to improve localization and classification accuracy.

\subsection{Weave Mamba Fusion (WMF)}
\label{sec:wmf}
WMF is the fusion operator applied at each cross-scale node of WeaveBiFPN.
Whereas BiFPN combines the two scales by weighted summation, WMF instead weaves
them into a single 2D map and processes it with a scan. It builds on the
SS2D~\cite{vmamba}, which extends the selective state space
model of Mamba~\cite{mamba}---itself based on structured SSMs~\cite{s4}---to 2D
data. SS2D scales linearly in the sequence length $L$ but quadratically in the
channel width $C$, giving $\mathcal{O}(L \cdot C^{2})$. This asymmetry motivates
WMF's two design choices: the linear cost in $L$ makes spatial weaving
inexpensive, while the quadratic cost in $C$ motivates applying the scan to only
a fraction of the channels.

Concretely, WMF weaves the two same-resolution maps column by column into a
single 2D map, so that every step of SS2D's horizontal scan crosses between the
two scales, realizing the global cross-scale interaction. Applying SS2D to only
a fraction of the channels then keeps this interaction efficient. Together,
these two choices let a single operator satisfy structure preservation, global
cross-scale interaction, and efficiency.

WMF operates on two inputs of the same spatial resolution. In the WeaveBiFPN
architecture, resolution alignment between adjacent levels is a prerequisite
for WMF: adjacent-level features are resized via bilinear upsampling
(top-down) or strided downsampling (bottom-up) to match the target level's
dimensions. We denote the resulting feature tensors of the current and
resized adjacent levels as $X_1, X_2 \in \mathbb{R}^{B \times C \times H \times W}$.

\noindent\textbf{(1) Width-axis Weaving}
$X_1$ and $X_2$ are interleaved column-wise to form a single 2D map
$Z \in \mathbb{R}^{B \times C \times H \times 2W}$:
\begin{equation}
Z[:,:,:,2i] = X_1[:,:,:,i],\quad
Z[:,:,:,2i+1] = X_2[:,:,:,i],
\quad i \in \{0, \ldots, W-1\}.
\end{equation}
Even columns of $Z$ thus come from $X_1$ and odd columns from $X_2$, so
the two scales alternate along the width.

\noindent\textbf{(2) Partial-channel Split}
We apply layer normalization to the woven feature map $Z \in \mathbb{R}^{B \times C \times H \times 2W}$. To optimize computational efficiency, we employ a partial-channel strategy that partitions the normalized features into two distinct streams. Specifically, the features are split along the channel dimension as follows:
\begin{equation}
\bar{Z} = \mathrm{LN}(Z), \qquad
(Z_{\mathrm{p}}, Z_{\mathrm{u}}) = \text{Split}(\bar{Z}), \qquad
C_{\mathrm{p}} = C/N,
\end{equation}
where $\bar{Z}$ is divided into a processed component $Z_{\mathrm{p}} \in \mathbb{R}^{B \times C_{\mathrm{p}} \times H \times 2W}$ and an identity component $Z_{\mathrm{u}} \in \mathbb{R}^{B \times (C - C_{\mathrm{p}}) \times H \times 2W}$. 
\begin{equation}
\hat{Z}_{\mathrm{p}} = \mathrm{SS2D}(Z_{\mathrm{p}}),
\qquad
\hat{Z}_{\mathrm{u}} = Z_{\mathrm{u}}.
\end{equation}
The SS2D operator is applied exclusively to $Z_{\mathrm{p}}$, while $Z_{\mathrm{u}}$ bypasses the scan and is passed forward via an identity mapping. This partial execution exploits channel redundancy, since only $1/N$ of the channels are needed to capture the essential cross-scale dependencies. The two streams are then recombined via a $1\times1$ convolution. We set $N{=}2$ by default, so that the scan is applied to half of the channels while the remaining half preserve the
original signal through the identity path.


As $Z_{\mathrm{p}}$ contains features from both scales, each horizontal scan step exchanges information across them. Because this exchange occurs along the width, a bidirectional horizontal scan ($s{=}2$) suffices, allowing us to remove the two vertical scans in VMamba's SS2D.

\noindent\textbf{(3) De-weaving}
The two branches are concatenated along the channel axis, then de-woven
column-wise to recover two scale-specific feature maps:
\begin{gather}
\hat{Z} = \mathrm{Concat}(\hat{Z}_{\mathrm{p}}, \hat{Z}_{\mathrm{u}})
\in \mathbb{R}^{B \times C \times H \times 2W}, \\
Y_1[:,:,:,j] = \hat{Z}[:,:,:,2j], \quad
Y_2[:,:,:,j] = \hat{Z}[:,:,:,2j+1], \quad j \in \{0, \ldots, W-1\}.
\end{gather}
As the exact inverse of the weave, de-weaving restores every column to its
source---even columns to $X_1$, odd columns to $X_2$---recovering the
original spatial structure of both feature maps without introducing any
parameters.

\noindent\textbf{(4) Channel Re-mixing}
The two outputs are concatenated along the channel axis, projected back to
$C$ channels, and passed through GELU:
\begin{equation}
Y = \mathrm{GELU}\!\left(\mathrm{Conv}_{1\times1}([Y_1;\, Y_2])\right).
\end{equation}
The $1{\times}1$ convolution fuses $Y_1$ and $Y_2$, allowing the
SS2D-processed channels to interact with the identity-branch channels so
that every channel contributes to the final representation.

\subsection{Complexity}
\label{sec:complexity}

Let $L = H \cdot W$ be the number of tokens in each input. We analyze the cost of a single application of WMF, since WeaveBiFPN
applies it at every cross-scale node. The cost is dominated by two operations: the SS2D scan and the final $1{\times}1$ convolution.

Applied to the full $C$ channels, SS2D's input and output projections would cost $\mathcal{O}(2L \cdot C^{2})$, where the factor of $2L$ comes from the woven width $2W$. The partial-channel design instead processes only $C/N$ channels, reducing this to $\mathcal{O}(2L \cdot C^{2}/N^{2})$---an $N^{2}\times$ reduction. The final $1{\times}1$ convolution then maps the $2C$ concatenated channels (SS2D-processed and identity branches) back to $C$ channels at width $W$, costing $\mathcal{O}(2L \cdot C^{2})$, where the factor of two now comes from the doubled input channels rather than the woven width. Weaving and de-weaving only rearrange elements and add no parameters, costing just $\mathcal{O}(L \cdot C)$.

Summing these terms, the $1{\times}1$ convolution dominates and the total
cost is $\mathcal{O}(L \cdot C^{2})$---linear in the number of tokens---so each WMF call remains efficient even at the high resolutions of WIDER FACE.

\subsection{WeaveBiFPN: Fully Bidirectional Pyramid with WMF}
\label{sec:weavebifpn}
The top-down pathway propagates semantic context by upsampling, which interpolates each high-resolution position from its nearest
low-resolution neighbors, smoothing out fine detail and producing
blurred, imprecise features (feature aliasing~\cite{fpn}). Because a
local fusion combines features only at nearby positions, it can only
partially correct this imprecision. WMF instead draws on spatially
distant context to correct it more fully. We therefore apply WMF at
every top-down node, fusing each level with the upsampled feature from above:
\begin{equation}
P_l^{td} = \mathrm{WMF}\!\left(\phi(\bar{w}_1\, P_l^{in}),\;
                               \phi(\bar{w}_2\, \mathrm{Up}(P_{l+1}^{td}))\right),
\label{eq:weavebifpn_td}
\end{equation}
where $P_l^{td}$ denotes the intermediate feature at level $l$ on the
top-down pathway, $\mathrm{Up}(\cdot)$ denotes bilinear upsampling,
$\phi(\cdot)$ is the Swish activation, and $\bar{w}_1$ and $\bar{w}_2$
are learnable scalar weights normalized as in BiFPN~\cite{bifpn}.
In the bottom-up pathway, downsampling compresses each local region of the
high-resolution map into a single low-resolution location. A small face that
occupies only a few pixels is thus reduced to a single position and becomes
hard to distinguish from the background. Local fusion alleviates this problem
only partially, as each fusion step aggregates a small neighborhood of
positions. WMF addresses this limitation by aggregating context from spatially
distant positions within a single fusion step. We therefore apply WMF
symmetrically at every bottom-up node, fusing each level with the downsampled
feature from below:
\begin{equation}
P_l^{out} = \mathrm{WMF}\!\left(\phi(\bar{w}_1'\, P_l^{td}),\;
                                \phi(\bar{w}_2'\, \mathrm{Down}(P_{l-1}^{out}))\right),
\label{eq:weavebifpn_bu}
\end{equation}
where $P_l^{out}$ denotes the output feature on the bottom-up pathway
and $\mathrm{Down}(\cdot)$ denotes parameter-free $3{\times}3$ max
pooling with stride 2. In BiFPN, each intermediate output node also receives a same-level
skip connection. We remove this skip connection so that every node fuses
exactly two inputs, as WMF requires. At the pyramid boundaries, we set $P_4^{td} = P_4^{in}$ and $P_2^{out} = P_2^{td}$.


\subsection{Auxiliary Modules and Detection Heads}
\label{sec:aux}

To form a complete WeaveFace, we combine WeaveBiFPN with widely used, lightweight components.

\textbf{CBAM Refinement.}
We apply CBAM~\cite{cbam} both before and after WeaveBiFPN, each with a residual connection over the input feature map $F$:
$F_{\mathrm{out}} = \mathrm{ReLU}(F + \mathrm{CBAM}(F))$. CBAM sequentially
combines channel and spatial attention. Pre-CBAM suppresses background noise
before it propagates into the SS2D scan, while Post-CBAM re-emphasizes channels
and regions already salient in the fused features.

\textbf{Detection Head.}
The detection head comprises a Cascade Context Prediction Module (CCPM)
followed by a ShuffleNet block~\cite{shufflenet}. To enlarge the per-scale
receptive field, we build on the three-branch CCPM of EResFD~\cite{eresfd}
but replace its standard convolutions with deformable convolutions
(DCNv3~\cite{dcnv3}), whose input-adaptive sampling supplies flexible
intra-scale spatial context that complements WMF's cross-scale global
fusion. The subsequent ShuffleNet block ($1{\times}1 \rightarrow$
depthwise $3{\times}3 \rightarrow 1{\times}1$, preceded by channel
shuffling) promotes cross-group mixing at minimal cost. Three parallel
$1{\times}1$ heads then produce classification scores, bounding-box
regressions, and landmark localizations per anchor. Predictions from all
pyramid levels are concatenated and supervised by the RetinaFace
multi-task loss~\cite{retinaface}:

\begin{equation}
\mathcal{L} = \mathcal{L}_{\mathrm{cls}}
            + \lambda_1 \mathcal{L}_{\mathrm{box}}
            + \lambda_2 \mathcal{L}_{\mathrm{ldm}},
            \qquad \lambda_1 = 2,\;\; \lambda_2 = 1,
\end{equation}
with softmax cross-entropy for $\mathcal{L}_{\mathrm{cls}}$ and smooth-$L_1$ for the box and landmark regression terms. At inference, softmax scores and non-maximum suppression yield the final detections.

\section{Experimental Results and Analysis}
\label{sec:experiments}

\subsection{Experimental Setup}

\subsubsection{Dataset.}
We evaluate WeaveFace on the WIDER FACE benchmark, which comprises 32,203 images and 393,703 annotated faces across three difficulty levels: Easy, Medium, and Hard. Since WIDER FACE lacks landmark ground truth, we train the landmark regression branch using the five-point annotations provided by RetinaFace~\cite{retinaface}.

\subsubsection{Implementation Details.}
\label{sec:implementation}
During training, input images are resized to $640\times640$ pixels. To improve generalization, we apply random square cropping, photometric distortion (brightness, contrast, saturation, and hue), horizontal flipping, zero-padding to a square shape, and mean subtraction. The model is optimized using AdamW with an initial learning rate of $1\!\times\!10^{-3}$ and a OneCycleLR schedule. We train the network for 450 epochs with a batch size of 32 on a single NVIDIA RTX A6000 GPU. To ensure a fair comparison with prior works, FLOPs are calculated at a standard VGA resolution of $640\times480$.
At inference, we adopt multi-scale testing, where input images are resized to multiple scales and the resulting predictions are merged via non-maximum suppression (NMS). The latency reported in Table~\ref{tab:abl-component} is averaged over 1000 runs at $640\times480$ resolution with batch size 1, following 50 warmup iterations, and measured with GPU--CPU synchronization enabled.

\begin{table}[h]
\centering
\renewcommand{\arraystretch}{1.15}
\caption{Comparison with state-of-the-art face detectors on the WIDER
FACE validation set (multi-scale AP). Fusion denotes the operator
family used for multi-scale feature fusion and context aggregation.
Results are from original publications or the multi-scale evaluations
in~\cite{scrfd} ($^\ddagger$) and~\cite{eresfd} ($^\S$). $^\dagger$:
self-attention baseline (ResNet-50 backbone with a DETR
Transformer encoder~\cite{detr}).}
\label{tab:wider_face}
\resizebox{\columnwidth}{!}{%
\begin{tabular}{l|c|c|c|ccc|c|c|c}
\toprule
Model & Backbone & Neck & Fusion & Easy & Medium & Hard & Avg.\,(\%) & \#Params\,(M) & GFLOPs \\
\midrule
DSFD$^\ddagger$~\cite{dsfd}       & ResNet-152 & FEM    & Conv. & 96.60 & 95.70 & 90.40 & 94.23 & 120.06 & 259.55 \\
TinaFace$^\ddagger$~\cite{tinaface} & ResNet-50 & FPN  & Conv. & 97.00 & 96.30 & 93.40 & 95.57 & 37.98 & 172.95 \\
TransEnc-R50$^\dagger$            & ResNet-50 & FPN    & Self-attention & 93.03 & 92.89 & 88.56 & 91.49 & 33.72 & 32.92 \\
RetinaFace$^\ddagger$~\cite{retinaface} & ResNet-50 & FPN & Conv. & 96.70 & 96.10 & 91.40 & 94.73 & 29.50 & 37.59 \\
\midrule
SCRFD-10GF~\cite{scrfd}           & Basic Res & PANet  & Conv. & 95.93 & 94.95 & 90.81 & 93.90 & 3.86 & 9.98 \\
FaceBoxes$^\ddagger$~\cite{faceboxes} & - & - & - & 85.90 & 81.60 & 55.70 & 74.40 & 1.01 & 0.28 \\
SCRFD-0.5GF~\cite{scrfd} & \makecell{Depth-wise \\ Conv} & PANet & Conv. & 92.71 & 91.45 & 86.23 & 90.13 & 0.57 & 0.51 \\
\midrule
RetinaFace Lite$^\ddagger$~\cite{retinaface} & MobileNet0.25 & FPN & Conv. & 91.40 & 89.20 & 82.50 & 87.70 & 0.44 & 0.80 \\
FDLite~\cite{fdlite}              & BLite & FPN    & Conv. & 92.31 & 89.91 & 82.30 & 88.17 & 0.24 & 0.94 \\
MTCNN$^\S$~\cite{mtcnn}           & - & -     & - & 85.10 & 82.00 & 60.70 & 75.93 & 0.12 & 0.01 \\
EResFD~\cite{eresfd}              & EResNet & SepFPN & Sep.\ Conv. & 89.02 & 87.96 & 80.41 & 85.80 & 0.09 & 0.30 \\
\midrule
\textbf{Ours} & \textbf{EResNet} & \textbf{WeaveBiFPN} & \textbf{SS2D} & \textbf{94.33} & \textbf{92.76} & \textbf{87.14} & \textbf{91.41} & \textbf{0.34} & \textbf{1.16} \\
\bottomrule
\end{tabular}%
}
\end{table}

\begin{table}[h]
\centering
\footnotesize
\caption{Comparison on AFW and PASCAL Faces without fine-tuning, following
the AP protocol of~\cite{facedetectionbell}. All models are trained solely on WIDER
FACE. \textbf{Bold}: our model; \underline{underline}: best result in
each column.}
\label{tab:cross_dataset}
\begin{tabular}{l| c| c| c}
\toprule
Method & AFW AP\,(\%) & PASCAL AP\,(\%) & \#Params \\
\midrule
MogFace~\cite{mogface} & 99.85 & \underline{99.32} & 85.26M \\
EfficientSRFace-L~\cite{efficientsrface} & \underline{99.94} & 98.84 & 18.84M \\
\midrule
SCRFD-0.5GF~\cite{scrfd} & 98.60 & 98.54 & 0.57M \\
SCRFD-1.0GF~\cite{scrfd} & 99.70 & 98.60 & 0.64M \\
SCRFD-2.5GF~\cite{scrfd} & 99.82 & 98.91 & 0.67M \\
FaceBoxes~\cite{faceboxes} & 98.91 & 96.30 & 1.01M \\
\midrule
\textbf{WeaveFace (Ours)} & \textbf{99.50} & \textbf{98.77} & \textbf{\underline{0.34M}} \\
\bottomrule
\end{tabular}
\end{table}

\subsection{Quantitative Results}
Table~\ref{tab:wider_face} compares WeaveFace with state-of-the-art face detectors on the WIDER FACE validation set. Our results demonstrate that WeaveFace achieves a superior balance between computational efficiency and detection accuracy, effectively redefining the performance frontier in the lightweight regime.
With only 0.34M parameters and 1.16 GFLOPs, WeaveFace achieves the highest AP across all difficulty levels among detectors in the sub-0.5M parameter range. It outperforms FDLite~\cite{fdlite} by 2.02\%/2.85\%/4.84\% on Easy/Medium/Hard, and surpasses EResFD~\cite{eresfd}, which shares the same EResNet backbone, by 6.73\% on Hard.
WeaveFace also offers a favorable accuracy-efficiency trade-off against heavier detectors. It surpasses SCRFD-0.5GF~\cite{scrfd} on every subset with $40\%$ fewer parameters, and trails SCRFD-10GF by only $3.67\%$ on Hard despite using $11\times$ fewer parameters and $9\times$ fewer GFLOPs. With nearly $100\times$ fewer parameters and over $28\times$ fewer GFLOPs, it nearly matches the self-attention-based TransEnc-R50 in mean AP (91.41\% vs.\ 91.49\%).

Table~\ref{tab:cross_dataset}
assesses robustness beyond WIDER FACE. Despite having the fewest
parameters among all compared methods, WeaveFace attains 99.50\% AP on
AFW and 98.77\% AP on PASCAL Faces. It outperforms
SCRFD-0.5GF~\cite{scrfd} on both datasets (99.50\% vs.\ 98.60\% on AFW,
98.77\% vs.\ 98.54\% on PASCAL) while using 40\% fewer parameters, and
trails the much heavier MogFace~\cite{mogface} (85.26M parameters) by only
0.35\% on AFW. These results indicate that WMF's global cross-scale
interaction generalizes to unseen data distributions rather than
overfitting to WIDER FACE-specific patterns.

\subsection{Qualitative Results}
\label{sec:qualitative}

\begin{figure}[t]
    \centering
    \includegraphics[width=1\linewidth]{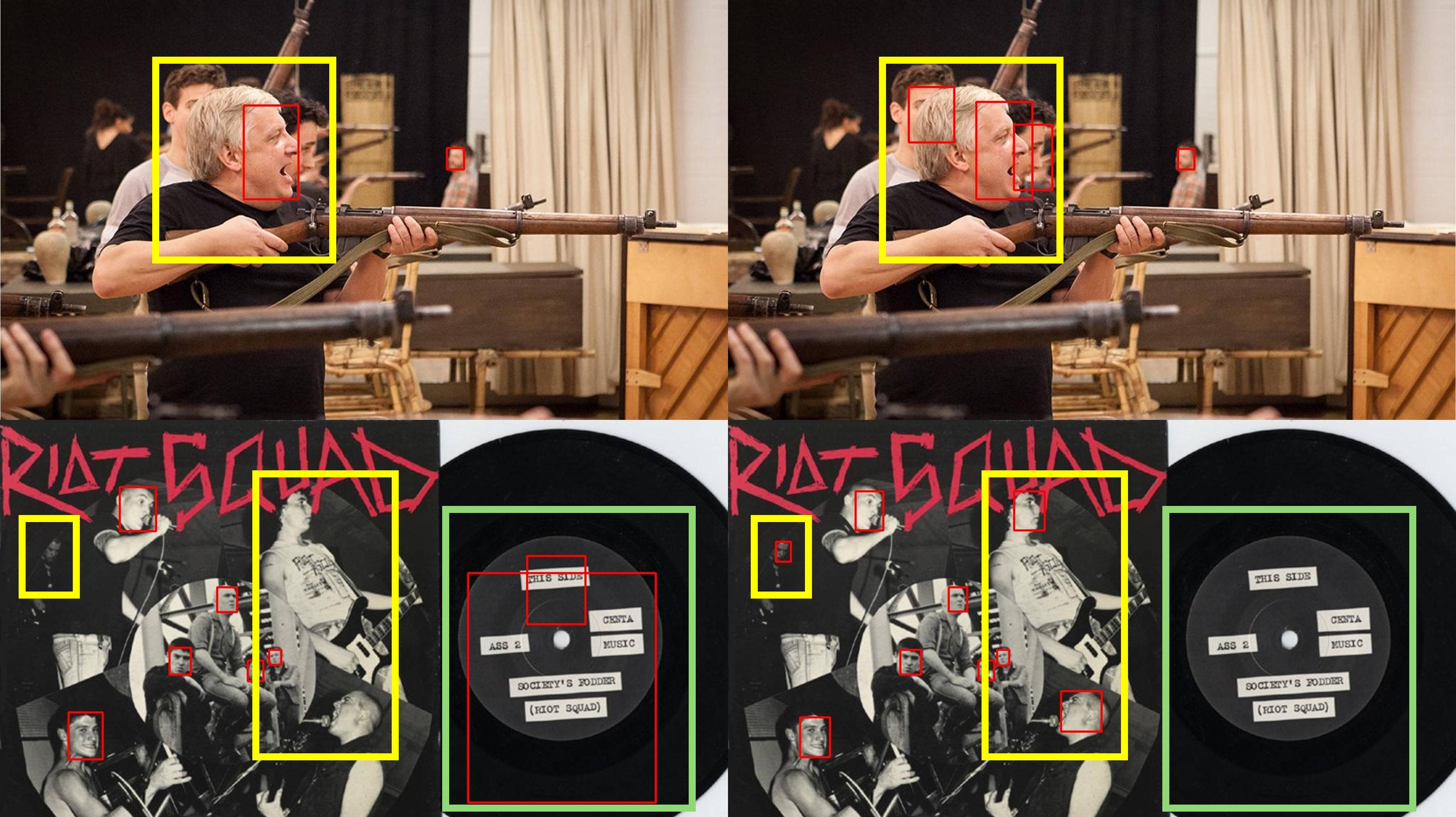}
    \caption{Qualitative comparison between EResFD (left) and WeaveFace (right) on the WIDER FACE Hard subset. Red boxes represent successful detections. Yellow boxes indicate faces missed by EResFD but correctly identified by WeaveFace, while green boxes highlight false positives generated by EResFD that are effectively mitigated by WeaveFace.}
    \label{fig:qualitative}
\end{figure}

Fig.~\ref{fig:qualitative} provides a qualitative comparison between EResFD and the proposed WeaveFace on challenging images from the WIDER FACE Hard subset. Both models are evaluated under identical conditions: a confidence threshold of 0.5, an NMS IoU threshold of 0.4, and single-scale inference at a fixed input resolution (short side of 1600 pixels). EResFD struggles in crowded scenes with small faces and heavy occlusion, producing both missed detections (yellow) and false positives (green). WeaveFace recovers most of these missed faces while suppressing the false positives, indicating improved robustness in dense and occluded scenes.


\subsection{Ablation Study}
\label{sec:ablation}
We conduct ablation experiments on the WIDER FACE~\cite{widerface} dataset
to evaluate the effectiveness of the proposed WMF: the contribution of the
weaving mechanism itself, its generality across pyramid topologies, the
design of the fusion operator together with the partial-channel divisor
$N$, and the effect of the scan count $s$. Unless otherwise specified, our default WeaveFace configuration applies WMF within a WeaveBiFPN, utilizing a partial-channel divisor of $N=2$ and a scan count of $s=2$; this configuration serves as the baseline for all ablation studies (Table~\ref{tab:wider_face}). All variants share the same EResNet backbone, CBAM attention, and CCPM detection head, and are trained under identical settings (Sec.~\ref{sec:implementation}). To ensure a rigorous evaluation, we isolate the component under study while keeping all other factors constant. 


\begin{table}[t]
\centering
\footnotesize
\caption{Comparison of different fusion operators on the BiFPN topology. We evaluate our proposed WMF against standard summation and cross-attention baselines. \textbf{Bold}: our model; \underline{underline}: best result in each column; \textbf{Lat.}: latency.}
\label{tab:abl-component}
\setlength{\tabcolsep}{2pt}
\begin{tabular}{lccccccc}
\toprule
Fusion & Easy & Medium & Hard & Avg. & \#Params\,(M) & FLOPs\,(G) & Lat.\,(ms) \\
\midrule
Sum (w/ sep. conv.)  & 92.88 & 90.87 & 84.76 & 89.50 & \underline{0.217} & \underline{0.564} & \underline{15.1} \\
Cross-attention      & 93.47 & 91.55 & 86.46 & 90.49 & 0.577 & 11.228 & 23.7 \\ \midrule
\textbf{WMF} & \textbf{\underline{94.33}} & \textbf{\underline{92.76}} & \textbf{\underline{87.14}} & \textbf{\underline{91.41}} & \textbf{0.344} & \textbf{1.159} & \textbf{23.3} \\
\bottomrule
\end{tabular}
\end{table}

\textbf{Effectiveness of the Weaving Mechanism}
\label{sec:abl-component}
We compare WMF against two standard fusion baselines on the BiFPN
topology (Table~\ref{tab:abl-component}). Simple summation with a
separable convolution reaches only $89.50$ mean AP, degrading sharply
on the Hard subset ($84.76$). Replacing summation with cross-attention
improves accuracy at every difficulty level ($90.49$ mean AP,
$+0.99$ over Sum), but at a steep efficiency cost: FLOPs rise by
nearly $20\times$ ($11.228$G vs.\ $0.564$G) and parameters more than
double ($0.577$M vs.\ $0.217$M).

Our proposed WMF attains the best accuracy on every subset, reaching
$91.41$ mean AP ($+1.91$ over Sum, $+0.92$ over cross-attention). Notably, WMF
achieves this while using only $0.344$M parameters --- roughly $40\%$
fewer than cross-attention --- and $1.159$G FLOPs, nearly an order of
magnitude ($9.7\times$) less compute. Despite this large FLOP
reduction, WMF's latency ($23.3$\,ms) is on par with, and in fact
marginally lower than, cross-attention ($23.7$\,ms), reflecting that
the selective scan underlying WMF is memory-bandwidth-bound rather
than compute-bound~\cite{mamba,mamba2}.


\begin{table}[t]
\centering
\footnotesize
\caption{Generality of WMF across pyramid topologies. Each topology is paired with its standard fusion method (summation for FPN/BiFPN, concatenation for PANet) and the proposed WMF. \textbf{Bold}: WMF (ours); \underline{underline}: best result in each column within each topology.}
\label{tab:abl-topology}
\setlength{\tabcolsep}{2pt}
\begin{tabular}{llcccccc}
\toprule
Topology & Fusion & Easy & Medium & Hard & Avg. & \#Params\,(M) & FLOPs\,(G) \\
\midrule
\multirow{2}{*}{FPN}
 & Sum   & 91.17 & 89.31 & 83.81 & 88.10 & 0.229 & 0.672 \\
 & \textbf{WMF} & \underline{91.47} & \underline{89.72} & \underline{84.12} & \underline{88.44} & \underline{0.187} & \underline{0.639} \\
\midrule
\multirow{2}{*}{PANet}
 & Concat      & 92.33 & 90.35 & 84.88 & 89.19 & 0.562 & 1.071 \\
 & \textbf{WMF} & \underline{92.75} & \underline{91.08} & \underline{85.03} & \underline{89.62} & \underline{0.329} & \underline{0.751} \\
\midrule
\multirow{2}{*}{BiFPN}
 & Sum   & 92.88 & 90.87 & 84.76 & 89.50 & \underline{0.217} & \underline{0.564} \\
 & \textbf{WMF} & \underline{94.33} & \underline{92.76} & \underline{87.14} & \underline{91.41} & 0.344 & 1.159 \\
\bottomrule
\end{tabular}
\end{table}

\textbf{Generality across Pyramids}
\label{sec:abl-topology}
Table~\ref{tab:abl-topology} replaces the standard fusion of FPN
(summation), PANet (concatenation), and BiFPN (summation) with WMF,
holding $N{=}2$, $s{=}2$ fixed across all three pyramids. 


WMF improves mean AP on every topology, and does so without added
cost on FPN and PANet: mean AP rises by $+0.34$ on FPN and $+0.43$ on
PANet while using 18--41\% fewer parameters and 5--30\% fewer FLOPs.
The gain is markedly larger on BiFPN ($+1.91$ mean AP), though here
WMF comes with more parameters ($0.217\text{M}\rightarrow0.344\text{M}$)
and FLOPs ($0.564\text{G}\rightarrow1.159\text{G}$). Adding parameters to the BiFPN fusion without weaving, however, yields a
far smaller gain, indicating that the improvement stems from the weaving
mechanism itself rather than the added capacity. BiFPN's bidirectional, weighted aggregation thus benefits the most from weaving, motivating our WeaveBiFPN neck.

\begin{table}[t]
\centering
\footnotesize
\caption{Accuracy–efficiency trade-offs across different scale-combination schemes (summation, concatenation, and weaving) as the partial-channel divisor $N$ increases. $N$ determines the extent of lightweighting in SS2D, where only $1/N$ of channels are processed; thus, a larger $N$ reduces model parameters and FLOPs at the expense of accuracy. \textbf{Bold}: our final configuration; \underline{underline}: best result in each column.}
\label{tab:abl-operator}
\setlength{\tabcolsep}{3pt}
\begin{tabular}{llcccccc}
\toprule
Fusion Operator & $N$ & Easy & Medium & Hard & Avg. & \#Params\,(M) & FLOPs\,(G) \\
\midrule
\multirow{3}{*}{Sum + SS2D}
 & 1 & 93.27 & 91.54 & 85.40 & 90.07 & 0.409 & 1.152 \\
 & 2 & 91.71 & 89.82 & 83.85 & 88.46 & 0.245 & 0.711 \\
 & 4 & 90.18 & 88.13 & 81.20 & 86.50 & \underline{0.193} & \underline{0.566} \\
\midrule
\multirow{3}{*}{Concat + SS2D}
 & 1 & \underline{95.14} & \underline{93.79} & \underline{88.40} & \underline{92.44} & 1.080 & 2.649 \\
 & 2 & 93.49 & 91.57 & 85.81 & 90.29 & 0.510 & 1.350 \\
 & 4 & 93.22 & 91.00 & 84.47 & 89.56 & 0.344 & 0.925 \\
\midrule
\multirow{3}{*}{Weave + SS2D (ours)}
 & 1 & 94.37 & 93.05 & 88.10 & 91.84 & 0.509 & 1.963 \\
 & \textbf{2} & \textbf{94.33} & \textbf{92.76} & \textbf{87.14}
                  & \textbf{91.41} & \textbf{0.344} & \textbf{1.159} \\
 & 4 & 93.17 & 91.57 & 86.20 & 90.31 & 0.292 & 0.870 \\
\bottomrule
\end{tabular}
\end{table}

\textbf{Fusion Operator and Partial-Channel Design}
\label{sec:abl-operator}
Table~\ref{tab:abl-operator} varies both the fusion operator (summation,
concatenation, weaving) and the partial-channel divisor $N$, which
controls how aggressively the channels are reduced for
lightweighting. At full channel width ($N{=}1$), concatenation attains
the highest mean AP ($92.44$), but only by roughly doubling the
parameters of weaving for a $+0.60$ gain.

This ranking reverses as channels are reduced. Weaving is markedly
more robust to channel reduction: halving the channels costs weaving
only $0.43$ AP, versus $2.15$ for concatenation and $1.61$ for
summation, and the gap persists at the $1/4$ ratio. The reason is
structural. Summation and concatenation merge the two scales into the
channel dimension before scanning, so reducing channels discards
information from both scales simultaneously. Weaving instead
interleaves the scales along the spatial width, so even a reduced
channel subset still carries a representation of both scales. As a
result, weaving outperforms concatenation at every reduced ratio while
using fewer parameters and FLOPs ($91.41$ vs.\ $90.29$ at $N{=}2$).

We adopt $N{=}2$ as our default: it comes within $1.03$ AP of
full-channel concatenation while using only a third of its parameters
and less than half of its FLOPs, making it the most favorable point on
the accuracy--efficiency trade-off.

\begin{table}[t]
\centering
\footnotesize
\caption{Effect of the scan count $s$ in WeaveBiFPN, where $s$ denotes the number of scan directions utilized by SS2D. \textbf{Bold}: our final configuration ($s{=}2$); \underline{underline}: best result in each column.}
\label{tab:abl-scan}
\setlength{\tabcolsep}{3pt}
\begin{tabular}{ccccccc}
\toprule
$s$ & Easy & Medium & Hard & Avg. & \#Params\,(M) & FLOPs\,(G) \\
\midrule
  1 & 93.87 & 92.07 & 86.51 & 90.82 & \underline{0.323} & \underline{0.991} \\
  \textbf{2} & \textbf{\underline{94.33}} & \textbf{\underline{92.76}} & \textbf{\underline{87.14}} & \textbf{\underline{91.41}} & \textbf{0.344} & \textbf{1.159} \\
  4 & 94.01 & 92.52 & 87.06 & 91.20 & 0.386 & 1.496 \\
\bottomrule
\end{tabular}
\end{table}

\textbf{Effect of Scan Count}
\label{sec:abl-scan}
With $N{=}2$ fixed, Table~\ref{tab:abl-scan} varies the scan count $s$
--- the number of scan directions, from forward horizontal ($s{=}1$)
to fully bidirectional along both axes ($s{=}4$). Adding the backward
horizontal pass ($s{=}1\rightarrow2$) reinforces the cross-scale
mixing already induced by weaving, improving mean AP. The further
vertical scans ($s{=}2\rightarrow4$) add only intra-scale context and
bring no additional gain. We therefore adopt $s{=}2$, reaching
$91.41$ mean AP at $0.344$M parameters and $1.159$G FLOPs.

\section{Conclusion}
In this paper, we studied multi-scale feature fusion for lightweight face
detection and identified the fusion operator itself as the core limitation
of pyramid-based detectors: combining two scales with summation or
concatenation, followed by local convolutions, only mixes features at
nearby positions and misses distant cross-scale context. To overcome this,
we proposed Weave Mamba Fusion (WMF), which weaves two adjacent feature
maps column by column and processes them with a 2D Selective Scan, so that
the scan mixes the two scales at every step and achieves global
cross-scale interaction. A partial-channel design applies SS2D to only
$1/N$ of the channels, cutting its dominant cost to roughly $1/N^{2}$ and
letting us place WMF at every node of a high-resolution pyramid. Placing
WMF in a fully bidirectional pyramid yields WeaveBiFPN, on which we build
WeaveFace. On WIDER FACE, WeaveFace attains a 91.41\% mean AP with only 0.34M
parameters and 1.16 GFLOPs, the most accurate detector with fewer than
0.5M parameters, with its largest gains on the Hard subset.

\textbf{Limitations}
The primary trade-off of WeaveFace lies in its latency. While WeaveFace achieves superior accuracy compared to summation-based baselines, it exhibits higher inference latency (23.3\,ms vs.\ 15.1\,ms). This is primarily attributed to the memory-bandwidth-bound nature of the selective scan operation~\cite{mamba,mamba2}, which limits the performance gains achievable solely through FLOPs reduction. We acknowledge that the current implementation does not fully exploit hardware parallelism. In future work, we aim to integrate the Mamba-2~\cite{mamba2} formulation, which reframes scanning as matrix multiplications, and further optimize inference by fusing the weaving step directly into the scan kernel to mitigate this overhead.

%
\bibliographystyle{splncs04}
\bibliography{main}

\clearpage
\appendix

\section{Generality across Backbones}
\label{sec:abl-backbone}
To investigate whether WMF is agnostic to the choice of backbone, we fix the WeaveBiFPN neck ($N{=}2$, $s{=}2$), CBAM, and CCPM head, while varying only the backbone architecture. Specifically, we evaluate two lightweight models (EResNet and MobileNetV1-0.25) and two heavy models (ResNet-50 and ResNet-152). All backbones are initialized with ImageNet-pretrained weights and trained under identical settings to the main model.

As shown in Table~\ref{tab:abl-backbone}, MobileNetV1-0.25 achieves slightly higher accuracy than EResNet; however, it incurs $1.6\times$ more parameters and violates the sub-0.5M budget constraint targeted in this work, making EResNet our default choice. When scaling up further, ResNet-152 more than doubles the parameter count of ResNet-50 without yielding performance gains, indicating that accuracy saturates around the capacity of ResNet-50 under our training recipe. Overall, these results demonstrate that the proposed WeaveBiFPN neck trains stably and scales consistently across backbones of varying capacities without requiring any backbone-specific tuning.

\begin{table}[h]
\centering
\caption{WeaveFace with different backbones on the WIDER FACE validation
set. The WeaveBiFPN neck ($N{=}2$, $s{=}2$), CBAM, and CCPM head are
fixed, and only the backbone is changed. The same fusion design
generalizes from lightweight to heavy backbones. \textbf{Bold}: our
default backbone.}
\label{tab:abl-backbone}
\setlength{\tabcolsep}{4pt}
\begin{tabular}{lcccccc}
\toprule
Backbone & Easy & Medium & Hard & Avg. & \#Params\,(M) & FLOPs\,(G) \\
\midrule
\textbf{EResNet} & \textbf{94.33} & \textbf{92.76} & \textbf{87.14}
                 & \textbf{91.41} & \textbf{0.344} & \textbf{1.159} \\
MobileNetV1-0.25 & 95.06 & 93.64 & 88.52 & 92.41 & 0.566 & 1.231 \\
ResNet-50        & 96.29 & 95.44 & 91.20 & 94.31 & 29.756 & 36.275 \\
ResNet-152       & 96.33 & 95.44 & 91.15 & 94.31 & 64.392 & 81.871 \\
\bottomrule
\end{tabular}
\end{table}





\section{Anchor Settings}
\label{sec:anchor}

As shown in Table~\ref{tab:anchor}, two anchors with a 1:1 aspect ratio
are assigned to every spatial location of the three pyramid levels
$P_2$--$P_4$, following the anchor design of
RetinaFace~\cite{retinaface}. The six anchor sizes double from 16 to 512 pixels, and the pixel sizes of the anchors remain fixed across input resolutions. For label assignment and hard negative mining, we follow the matching strategy and loss formulation of RetinaFace.

\begin{table}[htb!]
\centering
\footnotesize
\caption{Anchor settings of WeaveFace. Feature-map sizes are given for
the $640\times640$ training resolution.}
\label{tab:anchor}
\setlength{\tabcolsep}{6pt}
\begin{tabular}{lccc}
\toprule
Level & Stride & Feature map & Anchor sizes (px) \\
\midrule
$P_2$ & 8  & $80\times80$ & 16, 32 \\
$P_3$ & 16 & $40\times40$ & 64, 128 \\
$P_4$ & 32 & $20\times20$ & 256, 512 \\
\bottomrule
\end{tabular}
\end{table}



\section{Generality to Generic Object Detection}
\label{sec:abl-generic}

To verify that WMF is not limited to face detection, we integrate it
into the BiFPN neck of EfficientDet~\cite{bifpn} and evaluate on
two generic object detection benchmarks, COCO val2017 and PASCAL VOC.
The backbone, detection head, and training settings are kept identical
to the original EfficientDet, so the only difference between the two
variants is the addition of WMF. No face-specific components (\eg,
CBAM, CCPM, or landmark regression) are used.

\begin{table*}[t]
\centering
\caption{Results on COCO val2017 and PASCAL VOC datasets. Only WMF is added to the BiFPN neck of EfficientDet; all other settings are unchanged.}
\label{tab:abl-side-by-side}
\footnotesize
\setlength{\tabcolsep}{1pt}
\begin{tabular}{l ccccc | ccc}
\toprule
& \multicolumn{5}{c|}{\textbf{COCO val2017}} & \multicolumn{3}{c}{\textbf{PASCAL VOC}} \\
\cmidrule(lr){2-6} \cmidrule(lr){7-9}
\textbf{Variant} & \textbf{AP} & \textbf{AP$_{50}$} & \textbf{AP$_{75}$} & \textbf{\shortstack{Params\\(M)}} & \textbf{\shortstack{FLOPs\\(G)}} & \textbf{mAP} & \textbf{\shortstack{Params\\(M)}} & \textbf{\shortstack{FLOPs\\(G)}} \\
\midrule
EfficientDet (baseline)
  & 34.5 & 52.9 & 36.6 & 3.88 & 2.57
  & 76.90 & 3.84 & 2.35 \\
\textbf{EfficientDet + WMF}
  & \textbf{36.2} & \textbf{53.6} & \textbf{38.2} & 4.25 & 3.21
  & \textbf{78.63} & 4.21 & 2.99 \\
\bottomrule
\end{tabular}
\end{table*}

As reported in Table~\ref{tab:abl-side-by-side}, WMF improves the COCO AP from 34.5 to 36.2, yielding a performance gain of $+1.7$ points. Notably, the improvement is more pronounced at the stricter IoU threshold ($+1.6$ in $\text{AP}_{75}$) compared to the 
looser one ($+0.7$ in $\text{AP}_{50}$), suggesting that WMF primarily enhances localization precision rather than merely increasing the number of detections. Similarly, on the PASCAL VOC benchmark, WMF elevates mAP from 76.90 to 78.63 ($+1.73$). Across both tasks, this performance boost is achieved with a 
marginal overhead of approximately 0.37M parameters and 0.64--0.71 GFLOPs, where the slight parameter variation stems from the differing number of classes in the detection head. 
These findings demonstrate that the benefits of WMF generalize beyond face detection, successfully transferring to generic multi-class object detection.
\end{document}